\documentclass[sigconf,nonacm]{acmart}

\makeatletter
\def\@ACM@checkaffil{
    \if@ACM@instpresent\else
    \ClassWarningNoLine{\@classname}{No institution present for an affiliation}%
    \fi
    \if@ACM@citypresent\else
    \ClassWarningNoLine{\@classname}{No city present for an affiliation}%
    \fi
    \if@ACM@countrypresent\else
        \ClassWarningNoLine{\@classname}{No country present for an affiliation}%
    \fi
}
\makeatother

\usepackage{csquotes}
\usepackage{hyperref}
\usepackage{url}
\usepackage{booktabs}
\usepackage{amsfonts}
\usepackage{nicefrac}
\usepackage{microtype}
\usepackage{xcolor}
\usepackage{graphicx}
\usepackage{colortbl}
\usepackage{graphicx}
\usepackage{amsmath}

\usepackage{amssymb}
\usepackage{booktabs}
\usepackage{epsfig}
\usepackage{multirow}
\usepackage{makecell}
\usepackage{float}
\usepackage{subcaption}
\usepackage{pifont}
\usepackage{enumitem}

\AtBeginDocument{%
  }

\begin{document}
\title{MovieFactory: Automatic Movie Creation from Text using Large Generative Models for Language and Images}

\author{Junchen Zhu$^1$,\hspace{1em} Huan Yang$^2$,\hspace{1em} Huiguo He$^2$,\hspace{1em} Wenjing Wang$^2$,\hspace{1em} Zixi Tuo$^2$,\hspace{1em} \\Wen-Huang Cheng$^3$,\hspace{1em} Lianli Gao$^1$,\hspace{1em} Jingkuan Song$^1$,\hspace{1em} Jianlong Fu$^2$ }
\affiliation{%
 \vspace{0em}\institution{Center for Future Media, University of Electronic Science and Technology of China$^1$ \hspace{0.3em} \\Microsoft Research$^2$ \hspace{0.3em} National Taiwan University$^3$}
}
\affiliation{%
  \institution{junchen.zhu@hotmail.com \hspace{0.2em} \{huayan,\hspace{0.2em}v-huiguohe,\hspace{0.2em}v-wenjiwang,\hspace{0.2em}v-zixituo,\hspace{0.2em}jianf\}@microsoft.com}
}
\affiliation{%
  \institution{wenhuang@csie.ntu.edu.tw \hspace{0.2em} lianli.gao@uestc.edu.cn \hspace{0.2em} jingkuan.song@gmail.com}
}

\renewcommand{\shortauthors}{Zhu et al.}

\begin{abstract}
In this paper, we present MovieFactory, a powerful framework to generate cinematic-picture (3072$\times$1280), film-style (multi-scene), and multi-modality (sounding) movies on the demand of natural languages. As the first fully automated movie generation model to the best of our knowledge, our approach empowers users to create captivating movies with smooth transitions using simple text inputs, surpassing existing methods that produce soundless videos limited to a single scene of modest quality. To facilitate this distinctive functionality, we leverage ChatGPT to expand user-provided text into detailed sequential scripts for movie generation. Then we bring scripts to life visually and acoustically through vision generation and audio retrieval. To generate videos, we extend the capabilities of a pretrained text-to-image diffusion model through a two-stage process. Firstly, we employ spatial finetuning to bridge the gap between the pretrained image model and the new video dataset. Subsequently, we introduce temporal learning to capture object motion. In terms of audio, we leverage sophisticated retrieval models to select and align audio elements that correspond to the plot and visual content of the movie. Extensive experiments demonstrate that our MovieFactory produces movies with realistic visuals, diverse scenes, and seamlessly fitting audio, offering users a novel and immersive experience. Generated samples can be found in \href{https://www.youtube.com/watch?v=tvDknhMFhzk}{\color{blue}YouTube}/\href{https://www.bilibili.com/video/BV1qj411Q76P}{\color{blue}Bilibili} (1080P).
\end{abstract}



\maketitle

\section{Introduction}
\begin{displayquote}
    ``The cinema has everything in front of it, and no other medium has the same possibilities for getting it known quickly to the greatest number of people.''
    
    \hfill - by Cesare Zavattini
\end{displayquote}

Movies, considered one of the most esteemed artistic mediums, have enraptured audiences for well over a century. However, the allure of the silver screen comes hand in hand with substantial expenses, as the creation of exceptional films necessitates top-tier equipment and a considerable production team. For instance, the cinematic masterpiece "Avatar: The Way of Water" required a 12-year production process, accompanied by a budget estimated at around 400 million dollars. Despite the industry's persistent ambition to simplify the film production process, current techniques merely offer basic assistance in combining video and audio clips for editing purposes~\cite{filmmaker2002,Bootlegger2015}. Consequently, the notion of generating movies autonomously from scratch and empowering individuals in filmmaking continues to exist as an unattainable fantasy.

Automatic visual content generation has attracted considerable research attention over the years. Early methods, employing variational auto-encoders~\cite{VAE} or adversarial learning~\cite{GAN}, are limited in their ability to generate complex scenes. Leveraging diffusion models~\cite{DDPM,beatgan} and multi-model training, DALL-E 2~\cite{2022DALLE2} first achieves substantial advancements in open-domain text-to-image generation. To further mitigate the computational cost associated with diffusion models, Latent Diffusion~\cite{latentdiffusion} employs a variational auto-encoder to compress images into a down-sampled latent space. Building upon this, Stable Diffusion~\cite{stablediffusion} achieves notable performance by training with extensive data~\cite{laion5b}. Recent works also focus on improving text-to-image generation for high resolution~\cite{imagen}, language alignment~\cite{eDiff-I}, controllability~\cite{controlnet}, and customization options~\cite{DreamBooth}.

With the remarkable achievements in image generation~\cite{lab2pix,unifiedMM,Illustrator,deformable,texture_transformer}, researchers have ventured into the realm of video generation. A number of studies~\cite{CogVideo,magvit,nuwa,Phenaki} adopt Transformer-based architectures~\cite{transformer} to synthesize videos either in an autoregressive or non-autoregressive manner. Other approaches draw inspiration from image generation frameworks and utilize diffusion models. To mitigate the training cost associated with starting from scratch, text-to-video generation models often extend from pretrained text-to-image models~\cite{ImagenVideo,MakeAVideo,magicvideo,LVDM,VideoFusion}. One of the keys to video generation lies in establishing coherent connections across different frames. {Inherited from Stable Diffusion, Video LDM~\cite{VideoLDM} introduces temporal convolution and attention layers after each corresponding spatial layer, thereby addressing the out-of-distribution problem through fine-tuning each pre-trained model on videos.}

Despite significant advancements in video generation, achieving the desired standards of picture quality, audiovisual effects, and automation in generating \textbf{movies} remains a considerable challenge. Firstly, existing large-scale video datasets often exhibit subpar quality, which introduces artifacts like watermarks and hinders the model's adaptation to the cinematic ultrawide format. {Secondly, owing to the scarcity of research on the co-modeling of audio and video, current joint audio generation models~\cite{MMDiffusion,svgvqgan} fall short in producing content that meets satisfactory standards.} Lastly, current models lack the ability to adjust user-provided text inputs, which becomes particularly challenging when generating multiple scenes for a movie. Expecting users to provide detailed descriptions of sequential scripts is both unrealistic and user-unfriendly.

To tackle the above issues, we present MovieFactory, a text-to-movie framework capable of producing cinematic-picture (3072$\times$ 1280), film-style (multi-scene), multi-modality (sounding) movies, offering users a novel and immersive experience. First, to enable the automatic generation of multiple scenes, we leverage the capabilities of ChatGPT~\cite{chatgpt} to expand concise user descriptions into detailed scripts that collectively form the complete movie, with each script corresponding to a distinct scene. {Second, considering the inherent limitations entailed in generating audio content from scratch, we adopt an alternative approach by retrieving correspondingly aligned audio from a comprehensive database.} Third, to enhance the picture quality of generated videos, we propose a two-stage learning strategy comprising video-by-frame pretraining and video training. Adapting pretrained text-to-image models to the video domain presents challenges due to the visual domain shift between the pretrained image dataset and target video datasets. To overcome this, we employ fine-tuning on independent video frames, incorporating {domain-aware normalizations} and additional spatial layers to handle diverse spatial distributions across datasets, ensuring high-quality generation while mitigating the out-of-distribution problem. Subsequently, temporal layers are introduced and trained on videos following previous works~\cite{VideoLDM,CogVideo}. {Lastly, targeting high-quality movie production, we employ fine-tuning on a small collection of movie clips.}
Additionally, we incorporate a super-resolution model for better user experiences, referring to remarkable performances of existing works~\cite{realbasicvsr,spatiotemporal_frequency,trajectory_aware}. In summary, our contributions can be summarized as follows:
\begin{itemize}
    \item[1)]
    We propose MovieFactory, a movie generation framework that allows users to create high-definition (3072×1280), cinematic-style (ultrawide format), and multi-scene movies with accompanying sound by simply using text inputs.
    \item[2)]
    A two-stage training strategy is introduced to handle the visual domain shift between image and video datasets. Domain-aware normalizations and extra spatial layers enable the model to generate high-quality visual content even when trained on video datasets with limited quality.
    \item[3)] 
    We showcase the remarkable potential of combining large-scale AI models in the domain of automated movie generation, introducing a novel and promising application area for AI-generated content.
\end{itemize}

\begin{figure}[t]
    \centering
    \includegraphics[width=\linewidth]{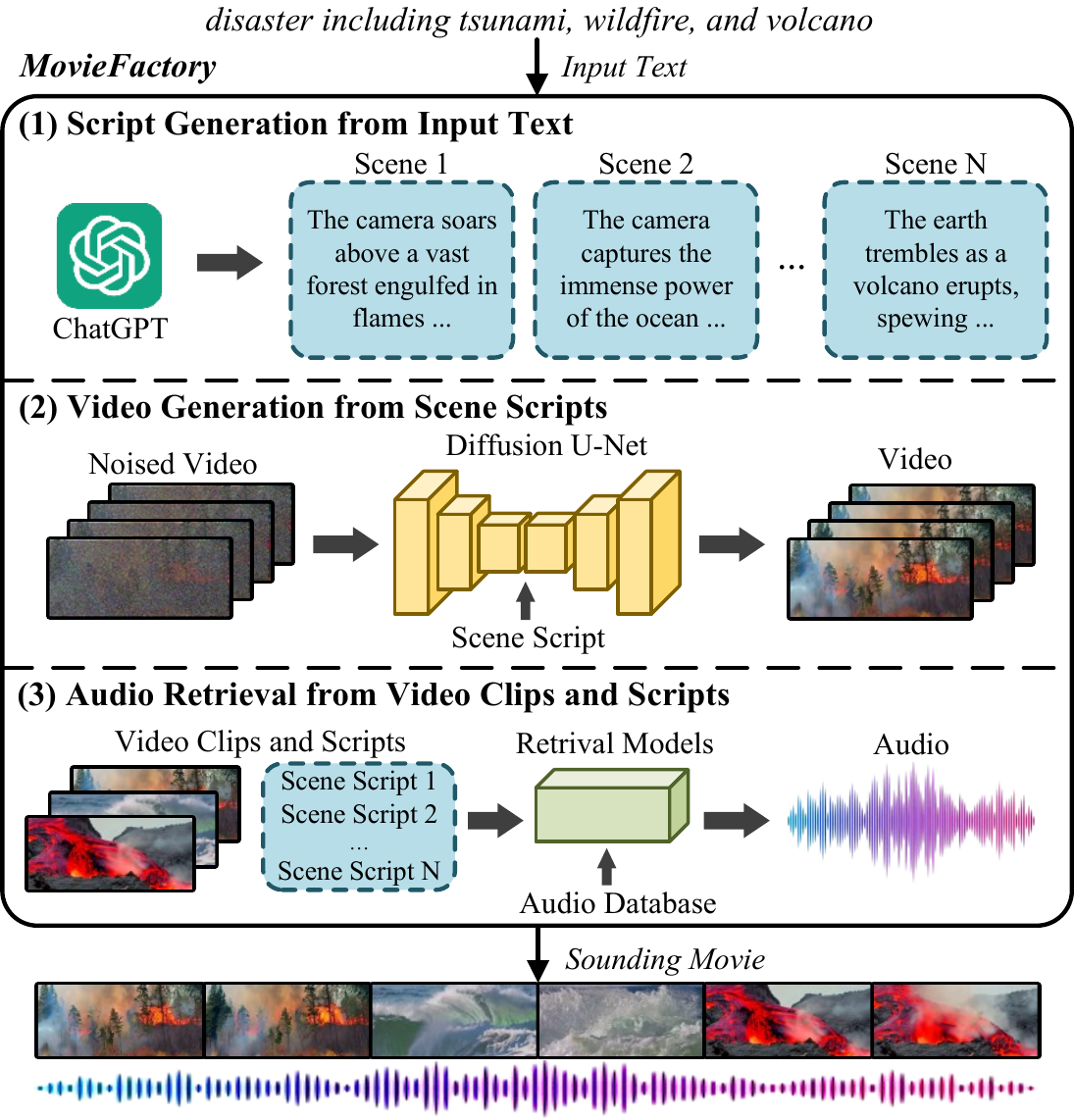}
    \caption{Illustration of our MovieFactory. Given an input text, we utilize ChatGPT to expand it into sequential detailed scripts, and each script describes one scene in the movie. Then, our model generates the visual content and retrieves the audio part for one movie clip with each script. Composing all clips, we obtain the final high-quality movie.}
    \label{fig:framework}
\end{figure}

\section{MovieFactory}
Movie generation goes beyond a mere combination of video and audio footage. It requires expertise in various domains, including scriptwriting, cinematography, directing, and sound effect design. Therefore, the creation of movies presents a greater challenge compared to basic video and audio generation, particularly when considering the development of a user-friendly interface for non-professionals. In this regard, we propose MovieFactory, a comprehensive and fully automated movie generation framework. With MovieFactory, users can effortlessly initiate the movie creation process by providing a concise description, which can be as simple as a single word indicating the main character.

\begin{figure*}[t]
    \centering
    \includegraphics[width=\linewidth]{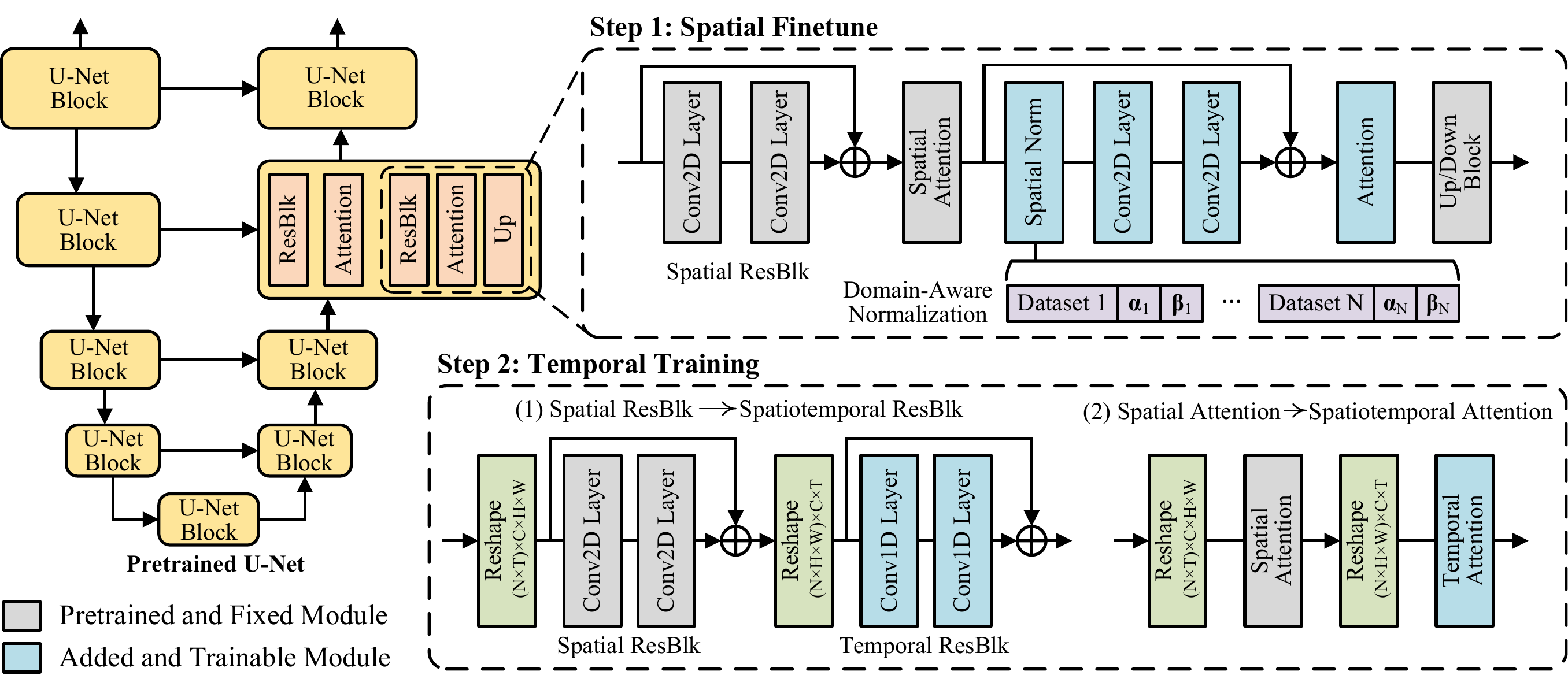}
    \caption{Demonstration of the two training stages of our video diffusion model. In each stage, we fix all components from the pretrained model, and only optimize the newly added blocks.}
    \label{fig:videomodel}
\end{figure*}

To fulfill the comprehensive requirement of movies, including high picture quality, smooth scene transitions, and video-audio synchronization, our design incorporates a range of components. The overall framework is shown in Fig.~\ref{fig:framework}. First, we leverage the power of a large language model to generate movie scripts of superior quality. Careful engineering of prompts ensures that the resulting movie plots adhere to fundamental principles of filmology and are well-suited for the subsequent audiovisual generation process. Second, we devise generation modules to bring each script to life visually and acoustically.
{Considering the restricted capabilities of current audio generation models, generating sound from scratch is not optimal. Therefore, we implement a two-stage process consisting of text-to-video generation and text\&video-to-audio retrieval. In the first stage, we use a diffusion model to construct videos by progressively removing noise from the input.  In the second stage, we retrieve synchronized audio from a comprehensive database that corresponds to the given context. Finally, we consolidate the generated sounding videos to obtain the complete movie.}

\subsection{Script Generation}
The input for our framework can be as simple as a concise plot description in a single sentence. However, considering the limitations of current video generation models that can only produce a single-scene clip from a solitary text prompt, it becomes essential to expand the user's input into a series of prompts, with each prompt describing a distinct scene.
These prompts collectively form a sequence of scripts.
We expect these scripts to adhere to the principles of scriptwriting while introducing innovative and unique perspectives to the subject matter. 
Furthermore, the prompts should effectively showcase the capabilities of video generation.

\makeatletter
\renewenvironment{quote}
               {\list{}{\listparindent=10pt
                        \leftmargin=12pt
                        \rightmargin=10pt
                        \topsep=0pt
                        \parsep        \z@ \@plus\p@}%
                \item\relax}
               {\endlist}
\makeatother

To attain this challenging objective, we leverage the powerful large language model, ChatGPT~\cite{chatgpt}, and integrate our requirements through prompt design. 
An example is shown below,
\begin{quote} \vspace{2pt}
\textbf{Prompt}: ``Write a sequence of prompts, using for movie generation for AI. Requirements:
1) each prompt only serves for one scene lasting for about 2 seconds;
2) each prompt contains clear subjects and detailed descriptions;
3) each prompt contains texts like "4K" and "high resolution" for leading high-quality generation;
4) the transition of each scene is very smooth;
5) no other character appears in this movie. The movie is about [User Input]''
\end{quote} \vspace{3pt}
\noindent By structuring prompts in this manner, we can guide the generation model to produce coherent and engaging movie contents.

To illustrate, let us create a movie about ``a race between a car and an airplane.'' Given our instruction, ChatGPT generates ten scripts, each corresponding to a distinct scene. In this ten-scene movie, the initial plot is on introducing the main characters through successive ``close-up'' shots of the ``airplane soaring through the sky,'' and the ``car speeding along a coastal road.'' As the movie progresses, captivating highlights unfold, including scenes such as ``car drifting around a hairpin turn,'' ``airplane diving through a narrow canyon,'' and ``car and airplane racing side by side.'' The movie concludes with a victorious moment captured in the scene titled ``checkered flag waving in victory.'' Each script for generation incorporates both the unfolding events and camera instructions, providing clear guidance for the generation process.

\subsection{Video Generation}
As indicated by prior studies~\cite{MakeAVideo,VideoLDM,CogVideo,Text2VideoZero,tuneavideo}, text-to-image pretraining plays a crucial role in open-domain text-to-video generation.
This is primarily due to the significant gap in scale and quality between current video datasets~\cite{webvid,hdvila} and well-established image datasets~\cite{laion5b}.
Following these works, we extend a pretrained image diffusion model to develop a video diffusion model.
We leverage the widely used Stable Diffusion\footnote{\url{https://github.com/Stability-AI/stablediffusion}}. To optimize our model, we incorporate two training steps: spatial finetuning and temporal training, as illustrated in Fig.~\ref{fig:videomodel}.

\subsubsection{Spatial Finetune} serves the purpose of bridging the spatial gap prior to capturing temporal information. In our approach, the Stable Diffusion model is pretrained on the LAION-5B dataset~\cite{laion5b}, which comprises high-quality images. Conversely, existing large-scale video datasets are limited in terms of resolution and visual quality, even containing watermarks.
Furthermore, our pretrained model is specifically optimized for generating square visual content, as it is trained on square images (height:width=1:1). Although minor adjustments in resolution have shown negligible effects on visual content and quality, significant changes in aspect ratio (e.g., transitioning from 1:1 to 2.35:1, as seen in movies) can lead to unstable generation, characterized by content ghosting and duplication. Therefore, it is crucial to address the spatial out-of-distribution discrepancy before delving into motion learning.

As Video LDM~\cite{VideoLDM} indicates, using low-quality video data to finetune the pretrained layers will inevitably harm the generation performance. Unfortunately, current large-scale dataset cannot satisfy good picture, motion, and text quality at the same time. In other words, if we finetune the whole model and fit the video distribution of the highest picture quality, the training may fail in the motion learning which is a disaster for the whole framework, and vice versa. To address above issues, we design a novel finetuning strategy to take advantage of different datasets as much as possible. Different from previous works which directly finetune the pretrained model, we fix the original model and insert extra layers to fit the distribution changes. There are two advantages of this design: 1) the whole knowledge in the pretraining can be completely remained, thus the contents and scenes that are not included in the video dataset can still be generated; 2) we can fit multiple distributions in the new modules, which solves the out-of-distribution problem in the next temporal training and keeps the ability to generate high-quality pictures at the same time. 
Specifically, as shown in Fig.~\ref{fig:videomodel}, we add a modified ResBlk and Attention layer before each Up or Down block in U-Net blocks. In the modified ResBlk, we add a learnable domain-aware normalization to specify and fit different spatial distributions. For each dataset, it learn a scaler $\alpha_{i}$ and shifter $\beta_{i}$, and works as follows:
\begin{equation}
    \textbf{H}=\textbf{X} \cdot \alpha_{i} + \beta_{i}
\end{equation}
where $\textbf{X}$ and $\textbf{H}$ are the input and output feature respectively, and $\alpha_{i}$ and $\beta_{i}$ are vectors with the same channel number as $\textbf{X}$ and $\textbf{H}$. For the structure of the added Conv2D and Attention layers, we completely follow previous blocks in the same U-Net Block. 

\subsubsection{Temporal Training} makes the model learn the motion of objects after the model is capable of generating images in the target distribution. Following previous works, we add the temporal layer after each pretrained spatial layer. Specifically, as illustrated in Fig.~\ref{fig:videomodel}, after each pretrained spatial ResBlk, we add a temporal ResBlk with 1D convolutions. Similarly, we add a temporal attention after each spatial attention, which shares the totally same hyper-parameter as the spatial one. Different from the pretrained spatial attention, following Video LDM~\cite{VideoLDM}, we also add sinusoidal embeddings~\cite{transformer} to the feature as the positional encoding for the time sequence.

\subsection{Audio Retrieval}
\begin{figure}[t]
    \centering
    \includegraphics[width=\linewidth]{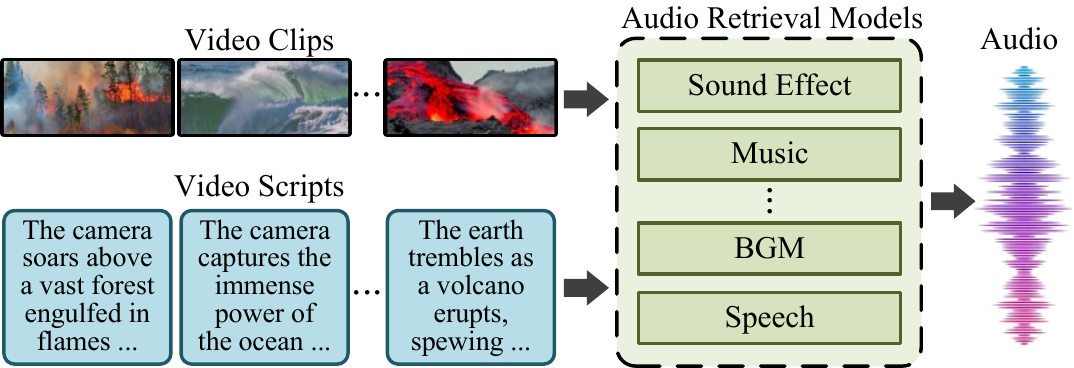}
    \caption{Illustration of our audio retrieval strategy.}
    \label{fig:audiomodel}
\end{figure}

Audio plays an indispensable role in movies.
By providing users with an additional sensory experience, audio has the capacity to enhance both the emotional impact and atmospheric quality of a scene.
Despite the self-evident importance of audio, limited research has been dedicated to the joint generation of video and audio. This can be attributed to challenges such as the scarcity of large-scale datasets and the limitations imposed by model size, leading to the current inability of existing models~\cite{MMDiffusion,svgvqgan} to generate high-quality audio. As an alternative, we adopted a retrieval-based approach, leveraging the richness of the audio database to align suitable sounds with the provided video and text.
Our audio retrieval strategy is illustrated in Fig.~\ref{fig:audiomodel}.

Sound in a movie includes music, dialogue, sound effects, ambient noise, and/or background noise~\cite{reich2017exploring}.
In this paper, we explore two distinct categories: sound effects and background music. 
Sound effects, such as footsteps, explosions, or door creaks, vary across different scenes to enhance realism and enrich the visual experience, immersing the audience in the on-screen actions. On the other hand, background music remains consistent throughout the entire movie sequence, serving to establish the overall tone, evoke emotions, and guide the audience's perception of the narrative. To address these two types, we propose different strategies.
For sound effects, we employ two retrieval approaches: text-to-audio~\cite{oncescu2021audio} and video-to-audio~\cite{suris2018cross}.
We extract features from the original scripts or the generated video content and match them with suitable audio clips from the database. 
As for background music, we leverage ChatGPT to summarize the plot and tone, and then combine the recommended tone category with techniques from music information retrieval~\cite{NasrullahZ19} to identify appropriate music tracks.

\begin{figure*}[t]
    \centering
    \includegraphics[width=\linewidth]{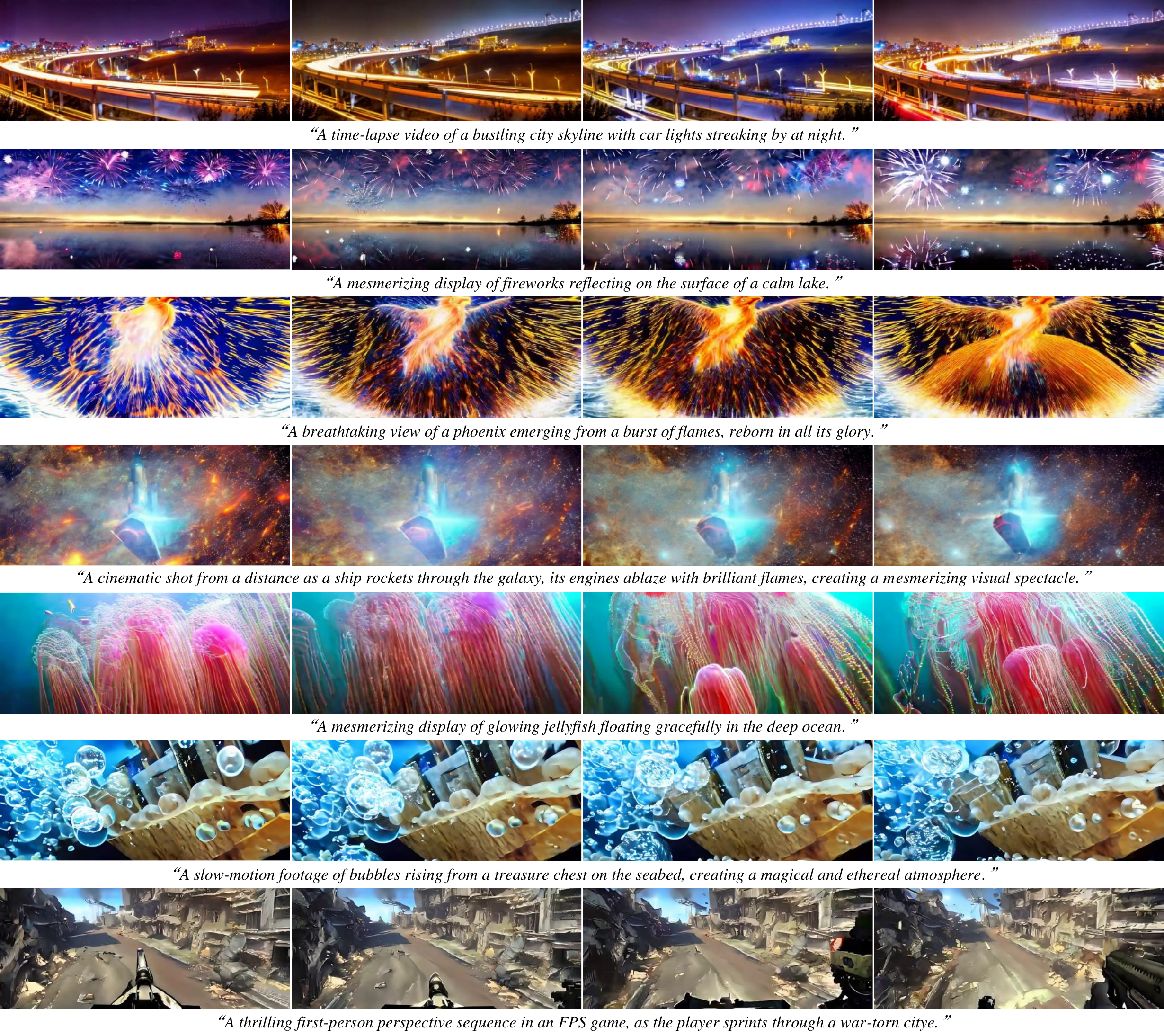}
    \caption{Generation video samples of our MovieFactory. Our model is able to generate both realistic and science fiction scenes in high quality, with rich details and smooth motion.
    More cases can be found in \href{https://www.youtube.com/watch?v=tvDknhMFhzk}{\color{blue}YouTube}/\href{https://www.bilibili.com/video/BV1qj411Q76P}{\color{blue}Bilibili}. Please play it in 1080p. All generated samples are for research purposes only and cannot be used for any commercial purposes.}
    \label{fig:videoresult}
\end{figure*}

\begin{figure*}[t]
    \centering
    \includegraphics[width=\linewidth]{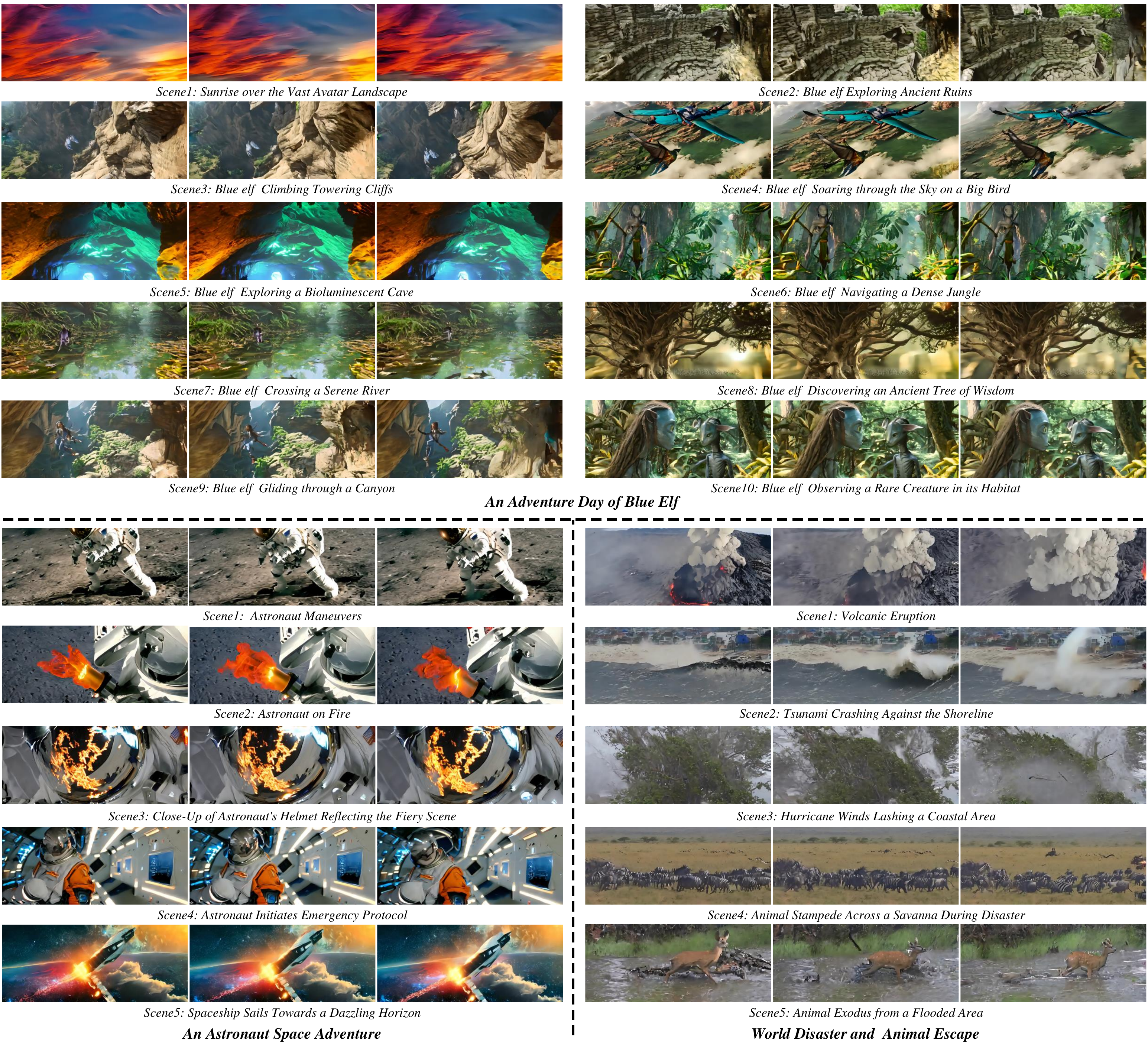}
    \caption{Generation movie samples of our MovieFactory. Given a subject of the movie, our model automatically generates the whole movie with multiple scenes.
    More cases can be found in \href{https://www.youtube.com/watch?v=tvDknhMFhzk}{\color{blue}YouTube}/\href{https://www.bilibili.com/video/BV1qj411Q76P}{\color{blue}Bilibili}. Please play it in 1080p and turn on the audio. All generated samples are for research purposes only and cannot be used for any commercial purposes.}
    \label{fig:movieresult}
\end{figure*}
\section{Experiments}
\subsection{Implementation Details}
We choose Stable Diffusion 2.0\footnote{\url{https://huggingface.co/stabilityai/stable-diffusion-2-base}} as the base image diffusion model, which is trained to synthesize 512$\times$512 images on LAION-5B~\cite{laion5b}. In the spatial finetune stage, we adopt WebVid-10M~\cite{webvid} and HD-VG-130M~\cite{VideoFactory} to jointly train the model in the scale of 768$\times$320. In the temporal training stage, we only use WebVid-10M to train the model for 16 frames generation with fps 8, where we use the normalization parameter learned for WebVid-10M in the spatial layers. Except for the resizing, random crop, and random flip, no other augmentation is involved in training. Also, no automatic or manual data filter is utilized in the pretraining. We adopt RealBasicVSR~\cite{realbasicvsr} to 4$\times$ upscale our generation results to obtain 3072$\times$1280 videos.

\subsection{Visual Generation}

Before evaluating the movie creation performance, we first assess the video generation capability of our model. In this section, we adopt the model that has been exclusively pretrained on the WebVid-10M dataset. We present the generated samples in Fig.~\ref{fig:videoresult}. The results demonstrate the effectiveness of our two-stage training strategy, as our model produces high-quality videos with clear visuals (without any watermarks) and smooth object motion. The generated videos exhibit rich details and showcase the successful application of our proposed approach.

For quantitative comparison, we utilize the Fréchet Video Distance (FVD)~\cite{FVD} to assess video quality and the CLIP similarity (CLIPSIM)~\cite{clip} for evaluating text-video alignment. We generate 5k samples using text extracted from the validation set of the WebVid-10M dataset. As shown in Tab.~\ref{tab:results_WebVid-15K}, compared with ModelScope~\cite{VideoFusion} and LVDM~\cite{LVDM}, our model performs better on both metrics.

\begin{table}[t]
    \centering
    \caption{Text-to-video generation on WebVid.}
    \vspace{-2mm}
    \begin{tabular}{>{\raggedright\arraybackslash}p{.3\linewidth} >{\centering\arraybackslash}p{.15\linewidth} >{\centering\arraybackslash}p{.15\linewidth}}
        \toprule
        Method           & FVD$\downarrow$ & CLIPSIM$\uparrow$ \\ 
        \midrule
        ModelScope~\cite{VideoFusion}        & 414.11 &  0.3000 \\
        LVDM~\cite{LVDM}     & 455.53 &  0.2751 \\ 
        \midrule
        Ours            & \textbf{317.52} & \textbf{0.3058} \\ 
        \bottomrule
    \end{tabular}
    \label{tab:results_WebVid-15K}
\end{table}

\subsection{Creating Movies}
Targeting the best visual quality as in movies, we finetune our model with some real processed movie clips after the pretraining. Note that, we only optimize the parameters of the spatial layers added in the stage of spatial finetune while fixing all other layers including all temporal ones. Fig.~\ref{fig:movieresult} demonstrates the impressive capabilities of our model in generating vivid and engaging movies.
For example, it can effectively depict the entire sequence of events experienced by astronauts in an emergency, as well as create captivating blue elf adventures that contain multiple scenes.

\section{Conclusion}
We introduce MovieFactory, a robust framework that revolutionizes movie generation. MovieFactory stands as the first model of its kind, enabling users to effortlessly create elaborate movies with cinematic-picture (3072$\times$1280), film-style (multi-scene), and multi-modality (sounding) using simple texts. To automatically generate multi-scene movies, we utilize ChatGPT to expand user descriptions into detailed scripts, each representing a distinct scene, thus forming a complete movie. To improve the visual quality of generated videos, we propose a two-step learning strategy involving video-by-frame pretraining and subsequent video training. Auxiliary spatial layers and domain-aware normalizations are applied to address the domain shift between pretrained image models and target video datasets, ensuring high-quality generation and mitigating out-of-distribution issues. Temporal layers are introduced and trained on videos to capture object motion and enhance the temporal coherence of the generated scenes. In order to produce high-quality movies, we fine-tune our models on a small collection of movie clips, further refining the generation process. Finally, rather than generating audio content from scratch, we employ a retrieval-based approach to align and retrieve corresponding audio from a comprehensive database. Extensive experiments validate that MovieFactory opens up a brand-new experience for users, empowering them to create captivating movies with ease and bringing a novel dimension to movie production.

\bibliographystyle{ACM-Reference-Format}
\bibliography{main}

\end{document}